\documentclass[10pt,twocolumn,letterpaper]{article}

\usepackage[pagenumbers]{cvpr} % To force page numbers, e.g. for an arXiv version

\usepackage{graphicx}
\usepackage{amsmath}
\usepackage{amssymb}
\usepackage{booktabs}
\usepackage{cite}
\usepackage{amsfonts}
\usepackage[accsupp]{axessibility}
\usepackage[ruled]{algorithm2e}
\usepackage{verbatim}
\usepackage{textcomp}
\usepackage{subcaption}
\usepackage{threeparttable}
\usepackage{url}
\usepackage[table,xcdraw]{xcolor}
\usepackage{colortbl}
\usepackage{booktabs}
\usepackage{wrapfig}
\usepackage{float}
\usepackage{multirow}
\usepackage{enumitem}
\usepackage{makecell}
\usepackage{setspace}

\usepackage{setspace}

\usepackage[pagebackref,breaklinks,colorlinks]{hyperref}

\usepackage[capitalize]{cleveref}
\crefname{section}{Sec.}{Secs.}
\Crefname{section}{Section}{Sections}
\Crefname{table}{Table}{Tables}
\crefname{table}{Tab.}{Tabs.}

\def\cvprPaperID{9030} % *** Enter the CVPR Paper ID here
\def\confName{CVPR}
\def\confYear{2023}

\begin{document}

\title{Orthogonal Annotation Benefits Barely-supervised Medical Image Segmentation}

\author{Heng Cai$^1$,~~Shumeng Li$^1$, Lei Qi$^2$, Qian Yu$^3$, Yinghuan Shi$^{1,}$\thanks{Corresponding author: Yinghuan Shi. Heng Cai, Shumeng Li, Yinghuan Shi and Yang Gao are with the State Key Laboratory for Novel Software Technology and National Institute of Healthcare Data Science, Nanjing University, China. This work is supported by the NSFC Program (62222604, 62206052, 62192783), CAAI-Huawei MindSpore (CAAIXSJLJJ-2021-042A), China Postdoctoral Science Foundation Project (2021M690609), Jiangsu Natural Science Foundation Project (BK20210224), and CCF-Lenovo Bule Ocean Research Fund.} , Yang Gao$^1$\\
$^1$Nanjing University~~~~~$^2$Southest University~~~~~$^3$Shandong Women's University\\
%{\tt\small echo@smail.nju.edu.cn}
}
\maketitle

%%%%%%%%% ABSTRACT
\begin{abstract}

Recent trends in semi-supervised learning have significantly boosted the performance of 3D semi-supervised medical image segmentation. Compared with 2D images, 3D medical volumes involve information from different directions, e.g., transverse, sagittal, and coronal planes, so as to naturally provide complementary views. These complementary views and the intrinsic similarity among adjacent 3D slices inspire us to develop a novel annotation way and its corresponding semi-supervised model for effective segmentation. Specifically, we firstly propose the orthogonal annotation by only labeling two orthogonal slices in a labeled volume, which significantly relieves the burden of annotation. Then, we perform registration to obtain the initial pseudo labels for sparsely labeled volumes. Subsequently, by introducing unlabeled volumes, we propose a dual-network paradigm named Dense-Sparse Co-training (DeSCO) that exploits dense pseudo labels in early stage and sparse labels in later stage and meanwhile forces consistent output of two networks. Experimental results on three benchmark datasets validated our effectiveness in performance and efficiency in annotation. For example, with only 10 annotated slices, our method reaches a Dice up to 86.93\% on KiTS19 dataset. Our code and models are available at \textcolor{magenta}{\url{https://github.com/HengCai-NJU/DeSCO}}.
\end{abstract}

%%%%%%%%% BODY TEXT
\begin{figure}[thb]
    \centering
    \vspace{-10pt}
    \includegraphics[width = 0.46\textwidth]{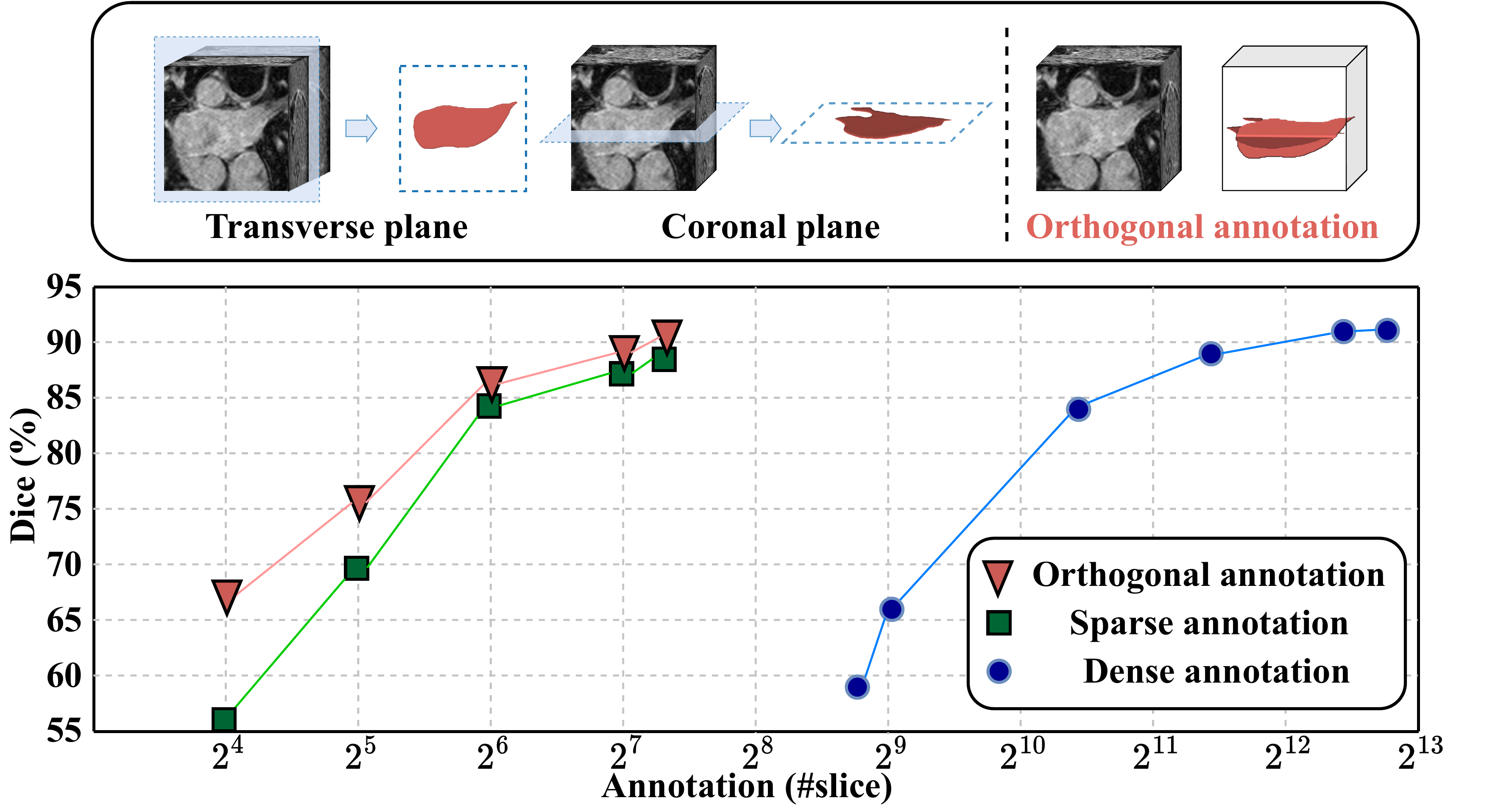}
    \caption{The upper figure illustrates our annotation method, each volume with annotations is labeled with only two orthogonal slices. The lower figure shows the comparison between the efficiency and effectiveness of our orthogonal annotation and other manners, including conventional dense annotation and previous sparse annotation which labels slices in one plane. All trained on LA~\cite{xiong2021global} dataset with supervised setting. For sparse annotation and our orthogonal annotation, we train the models only on labeled voxels through partial cross-entropy and partial Dice loss.} 
 \label{fig:annotation}
 \vspace{-10pt}
\end{figure}

\section{Introduction}
\label{sec:intro}
Medical image segmentation is one of the most critical vision tasks in medical image analysis field. Thanks to the development of deep learning-based methods~\cite{ronneberger2015u,cciccek20163d,milletari2016v,chen2017deeplab}, segmentation performance has now been substantially improved. 
However, the current promising performance is at the cost of large-scale manually precisely labeled dataset, which is prohibitively expensive and laborious to achieve. What's worse, different radiologists might provide different annotations even for a same image. Therefore, exploring ways to alleviate the requirement of quantity or quality of manual annotation is highly demanded. 
Mainstream methods typically follow two paradigms: 1) degrade annotation quality, \ie, weakly-supervised segmentation, and 2) reduce annotation quantity, \ie, semi-supervised segmentation.

Weakly-supervised segmentation methods usually utilize weak annotations, \eg, image-level label~\cite{lee2019ficklenet,lee2021anti}, scribble~\cite{lin2016scribblesup,liu2022weakly}, point~\cite{bearman2016s} or partial slices~\cite{bitarafan20203d,li2022pln}. Unfortunately, most of them are either difficult to distinguish some fuzzy boundaries or with additional large computational burden~\cite{jo2022recurseed}. What's more, weakly-supervised setting usually requires coarse annotation for every single image. This is still a heavy burden for radiologists. Besides, most current methods originally developed for 2D segmentation could not directly utilize 3D volumetric information~\cite{chen2022scribble2d5}.

Different from these weakly-supervised methods, semi-supervised methods train segmentation models with a small amount of manually labeled data and a large amount of unlabeled data, which have achieved remarkable performance with an impressive deduction on demand for annotation~\cite{li2018semi,bortsova2019semi}.
Despite their success, we notice most current semi-supervised segmentation methods still require full 3D annotation for each labeled volume. In fact, segmentation targets in adjacent slices of 3D volume are highly similar in both appearance and location, leading it redundant to label every slice. Although the sparse annotation is discussed in recent work~\cite{li2022pln}, we notice these conventional methods still neglect the complementary views between different directions in 3D volume. 

%Recent trends in semi-supervised learning ~\cite{xia20203d}~\cite{xia2020uncertainty}~\cite{chen2019learning} have revealed that learning from complementary view is indeed beneficial. It is known that 3D medical volumes naturally contains different directions (\eg, transverse, coronal planes) to form the learning paradigm from complementary views.Thus, we wonder \textit{whether a novel annotation method coupled with its corresponding model could be investigated by introducing this complementary relation into 3D semi-supervised medical image segmentation}.
It is known that 3D medical volumes naturally contains different directions (\eg, transverse, coronal planes) which provide complementary information from different views. And recent trends in semi-supervised learning ~\cite{xia2020uncertainty,chen2019learning} have revealed that learning from complementary view is indeed beneficial. Thus, we wonder \textit{whether a novel annotation method coupled with its corresponding model could be investigated by introducing this complementary relation into 3D semi-supervised medical image segmentation}.

%Despite their great success~\cite{li2018semi}\cite{bortsova2019semi}\cite{li2020transformation} with an impressive deduction on demand for annotation, the amount of labeled data (typically 20\% of the whole dataset) is still hard to fully ignore. 
%We notice that 1) objects in adjacent slices are highly similar in both shape and position, which makes it redundant to give a label for every slice; 2) Though organs may vary slightly from patient to patient, they still share some similarities in general location, shape, features, \etc, which makes it feasible to use only part of volumes instead of all of them to learn generalization ability. Based on the two facts, we propose our new method of delineation to further mitigate the need for annotation, that is to label few slices for small number of volumes in the dataset. Following~\cite{li2022pln}, we name the setting as Barely-Supervised Segmentation.

In this paper, for labeled volume, we innovatively investigate a novel sparse annotation way---\textit{orthogonal annotation}, \ie, only to label two slices in its orthogonal direction (\eg, transverse and coronal direction in Figure~\ref{fig:annotation}). We believe our annotation way has two merits: 1) it could largely force the model to learn from complementary views with two diversely initialized labeled slices, 2) it helps greatly reduce the label costs with fully utilizing the inter-slice similarity. Following very recent work~\cite{li2022pln}, we name the setting as Barely-supervised Segmentation.

%As our analysis in Sec.~\ref{subsec:orthoanno} shows, the proposed orthogonal annotation has its merits in disparity preservation, feature representation and performance promotion. The illustration of our annotation method is shown in Figure~\ref{fig:annotation}. According to calculation, our orthogonal annotation is highly efficient in labeling. For example, in LA~\cite{xiong2021global} dataset, our way only require to label 10 slices. 

%An intuitive thought to leverage the orthogonal annotation is to train a segmentation model where only the voxels on the labeled slices contribute to the training. However, directly learning from scarce sparse annotation is unstable and the training is apt to collapse (will be shown in Sec.~\ref{sec:Experiments}). To solve this problem, registration is a feasible and convenient solution to spread supervision signals from slice to volume and the result of label propagation can serve as the dense pseudo label for training. So through registration in orthogonal directions, it generates two sets of pseudo labels for volumes with orthogonal annotation. Still, the registration pseudo labels are not good enough to directly train a segmentation model using current existing semi-supervised methods. It is mainly due to the accumulation of error in the registration process. 

%From methodology part, an intuitive thought to incorporate our orthogonal annotation is to train a segmentation model where only the voxels on the labeled slices contribute to the training. 
To incorporate our orthogonal annotation, the most intuitive thought about training strategy of a segmentation model is that only the voxels on the labeled slices contribute to the training.
However, directly learning from this sparse annotation is unstable and the training is apt to collapse (shown in Sec.~\ref{sec:Experiments}). Thus, we apply registration to spread supervision signals from slice to volume, where the result of label propagation can serve as the dense pseudo label for training. By performing registration, we obtain two sets of pseudo labels for volumes from orthogonal directions. Yet, the obtained pseudo labels are not promising enough to directly train a segmentation model using current existing semi-supervised methods, which is mainly due to the accumulation of error in the registration process. 

Therefore, to leverage 1) the volumes with inaccurate pseudo labels and 2) the rest unlabeled volumes, we propose a simple yet effective end-to-end framework namely Dense-Sparse Co-training (DeSCO), which consists two segmentation models of a same structure. At the beginning of training, the models mainly learn from dense pseudo labels with a learning preference on voxels with more confident pseudo labels, \ie, voxels near to registration source slice, and exploit unlabeled volumes through cross-supervision. After the models have been improved through training, we gradually get rid of pseudo label until the supervised loss solely comes from sparse annotation. Meanwhile, the role of cross-supervision is gradually emphasized correspondingly. Because in the process of reaching consensus through cross-supervision, the mistake introduced by previous training on inaccurate pseudo labels could be revised.
Overall, our contributions are three folds: 

%1) \textbf{A novel barely-supervised medical image segmentation framework} to steadily utilize our high-efficient sparse annotation with coupled segmentation method. 2) \textbf{A new annotation way} that only labels two orthogonal slices for a labeled 3D volumetric image, which greatly reduce the annotation burden. 3) \textbf{A dense-sparse co-training paradigm} to learn from dense pseudo label and sparse label while leveraging unlabeled volumes to reduce noise by reaching consensus through cross-supervision.
\begin{itemize}
    \item A new annotation way that only labels two orthogonal slices for a labeled 3D volume, which greatly reduces the annotation burden.
    \item A novel barely-supervised 3D medical image segmentation framework to steadily utilize our high-efficient sparse annotation with coupled segmentation method. 
    \item A dense-sparse co-training paradigm to learn from dense pseudo label and sparse label while leveraging unlabeled volumes to reduce noise by reaching consensus through cross-supervision.
\end{itemize}

Extensive experiments on three public datasets validate that our barely-supervised method is close to or even better than its upper bound, \ie, semi-supervised methods with fully annotated labeled volumes. For example, on KiTS19, compared to Mean Teacher~\cite{tarvainen2017mean} that uses 320 labeled slices with a Dice of 84.98\%, we only uses 10 labeled slices yet obtains a Dice of 86.93\%. 

%our method with 10 labeled slices and MT~\cite{tarvainen2017mean} with 320 labeled slices reach a Dice coefficient of 86.93\% and 84.98\% respectively on KiTS19 dataset.

\begin{figure*}[thb]
    \centering
    \vspace{-10pt}
    \includegraphics[width = 0.95\textwidth]{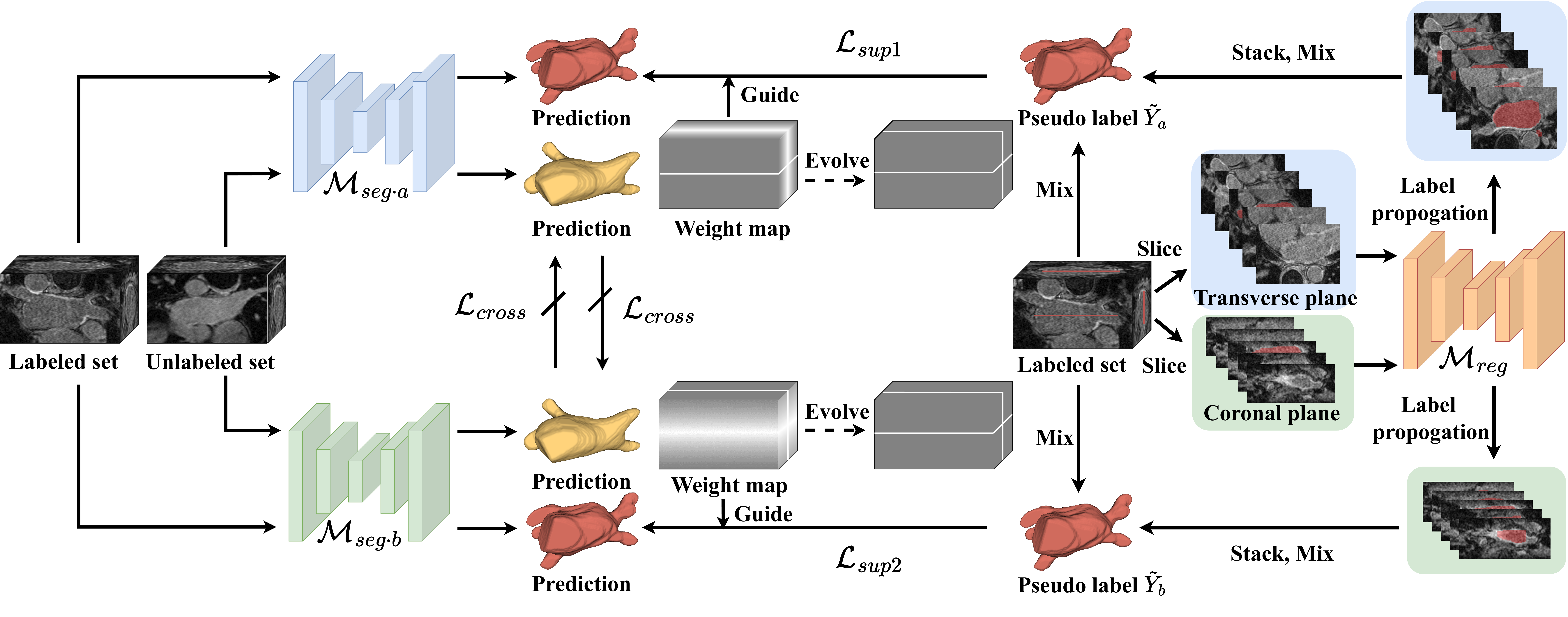}
    \caption{Overview of the proposed DeSCO paradigm. For a volume with orthogonal annotation, $\mathcal{M}_{reg}$ propagates the orthogonal annotation into the whole volume in two directions, and the results serve as pseudo labels to supervise segmentation model $\mathcal{M}_{seg\cdot a}$ and $\mathcal{M}_{seg\cdot b}$, respectively. For an unlabeled volume, $\mathcal{M}_{seg\cdot a}$ and $\mathcal{M}_{seg\cdot b}$ supervise each other with their outputs. Weight map guides $\mathcal{L}_{sup}$ and whiter areas mean higher weight.}
    \label{fig:framework}
    \vspace{-10pt}
\end{figure*}

\section{Related Work}
\textbf{Semi-supervised Learning.}
Semi-supervised learning methods leverage a few labeled data and abundant unlabeled data to reduce the dependency on laborious annotation, and show their effectiveness on various tasks~\cite{zhou2021towards,mi2022active,liu2022perturbed}. For example, $\Pi$-model~\cite{samuli2017temporal} minimized the inconsistency between the outputs of differently disturbed inputs in order to improve the consistency of the model. Temporal Ensembling~\cite{samuli2017temporal} and Mean Teacher~\cite{tarvainen2017mean} improved $\Pi$-model by leveraging the average of historic outputs to produce pseudo labels of higher quality. FixMatch~\cite{sohn2020fixmatch} and FlexMatch~\cite{zhang2021flexmatch} selected pseudo labels of higher quality with threshold. Some methods explored uncertainty by using novel consistency regularization, \eg, UPS~\cite{rizve2021defense} and NP-Match~\cite{wang2022np}. The consistency regularization way used in these methods has greatly inspired current mainstream semi-supervised medical image segmentation methods.

\textbf{Semi-supervised Medical Image Segmentation.}
To free radiologists from the burden of heavy annotation, many deep learning-based semi-supervised methods have been proposed for medical image segmentation in recent years. 

Consistency regularization is a popular method to leverage unlabeled data. Li \etal~\cite{li2018semi} made a consistency constraint between the outputs of original images and transformed images. Similarly, Bortsova \etal~\cite{bortsova2019semi} proposed a method that learns to predict consistently under different transformations. Xie \etal~\cite{xie2020pairwise} enforced the consistency of images in feature space. Different from previous work, Luo \etal~\cite{luo2021semi2} imposed the consistency constraint at task level. Shi \etal~\cite{shi2021inconsistency} incorporated consistency regularization and entropy minimization by treating certain/uncertain region with different training strategies. Besides, Co-training is another widely-used framework for semi-supervised medical image segmentation. Zhou \etal~\cite{zhou2019semi} applied multi-planar fusion to generate more accurate 3D pseudo label to train a co-training model. Xia \etal~\cite{xia2020uncertainty} introduced multi-view co-training into semi-supervised medical image segmentation and domain adaptation.  And some works~\cite{bai2017semi,yu2019uncertainty} improving the quality of pseudo labels also showed great effectiveness.
However, when applied to 3D segmentation tasks, as a heavy burden, most of them still require a dozen pixel-wise fully annotation slice-by-slice.

\textbf{Weakly-supervised Segmentation.}
 Weakly-supervised segmentation methods give pixel-level output with coarse annotation including image-level annotation~\cite{lee2019ficklenet,wang2020self,lee2021anti}, bounding box~\cite{dai2015boxsup,oh2021background}, scribble~\cite{lin2016scribblesup,liu2022weakly,wang2019boundary} and even some points~\cite{bearman2016s}. Among them, image-level annotation is the most popular setting and has been extensively studied. Many image-level weakly-supervised segmentation methods are based on class activation map (CAM). However, the area suggested by CAM is incapable of directly training a segmentation model as it only covers the most discriminative part of large objects and is liable to over-covers small objects. Other annotation methods suffer from the same problem more or less. The common problem in these annotation methods is that they discard the information of the boundaries between objects, which is crucial for segmentation tasks. Our annotation method takes into consideration the precise delineation of objects and utilizes volumetric information of 3D medical images.

\textbf{Remark.} Our setting could be regarded as a variant of semi-supervised segmentation by requiring much fewer annotated slices. Our proposed orthogonal annotation has its merits in preserving disparity to boost the performance in semi-supervised scenario. Also, compared with weakly-supervised segmentation, our method only requires very few annotated slices while yielding promising results.
\begin{comment}
\begin{figure}[thp]
\centering
    \begin{minipage}[t]{0.22\textwidth}
        \centering
        \includegraphics[width = 1\textwidth]{img/barchart3.pdf}
        \vspace{-10pt}
        \caption{HSIC of parallel slices and orthogonal slices.}
        \label{fig:hsic}
    \end{minipage}
    \begin{minipage}[t]{0.25\textwidth}
        \centering
        \includegraphics[width = 1\textwidth]{img/barchart4.pdf}
        \vspace{-10pt}
        \caption{Dice coefficient on test set of models trained with different annotations.}
        \label{fig:laeffe}
        \end{minipage}
\end{figure}

\begin{figure}[thb]
\vspace{-10pt}
    \centering
    \includegraphics[width = 0.43\textwidth]{img/tsne.pdf}
    \vspace{-1pt}
    \caption{t-SNE result of feature for classification.}
    \label{fig:tsne}
\end{figure}
\end{comment}

\begin{comment}
\begin{figure}[thb]
 \vspace{-10pt}
 \centering  
 %\subfigure[model $t_1$ and $t_2$]{
  \begin{subfigure}[b]{0.22\textwidth}
   
    \includegraphics[width=\linewidth]{img/tsnepal.png}
    \caption{model $t_1$ and $t_2$}
  \label{fig:tsnepal}
  \end{subfigure}
  \vspace{-10pt} 
  \hspace{-10pt} 
  \hfill
 %\subfigure[model $t_1$ and $c$]{
 \begin{subfigure}[b]{0.22\textwidth}
    
   \includegraphics[width=\linewidth]{img/tsnever.png}
   \caption{model $t_1$ and $c$}
  \label{fig:tsnever}
 \end{subfigure}
 \caption{t-SNE result of feature for classification.}
 \label{fig:tsne}
\end{figure}
\end{comment}

\section{Method}
In this work, we propose a novel annotation method called \textit{orthogonal annotation} and a coupled training method comprised of 1) a registration module and 2) DesCO segmentation model. The training procedure is illustrated in Figure~\ref{fig:framework}. We provide detailed introductions to the problem setting, our orthogonal annotation, registration module and the proposed DeSCO paradigm in the following parts.

\subsection{Problem Setting and Notations}
We aim to apply our orthogonal annotation in barely-supervised segmentation setting, \eg, only 10 orthogonal slices from 5 volumes for an entire 3D medical image dataset. Training set contains $N$ volumes $\{X_1,X_2,...,X_N\}$ of shape $H\times W\times D$, where $H, W$ and $D$ represent the height, width and depth of each volume, respectively. These volumes are further divided into $\{X_1,X_2,...,X_l\}$ with annotations and $\{X_{l+1},X_{l+2},...,X_N\}$ without annotations. Our methods annotates two slices from orthogonal planes denoted by $a$ and $b$. The labeled slices of $X_i (1\le i\le l)$ are denoted as $X_{ia}^m$ and $X_{ib}^n$, which means $m^{th}$ slice in plane $a$ and $n^{th}$ slice in plane $b$. Their annotations are denoted as $Y_{ia}^m$ and $Y_{ib}^n$.

\subsection{Justification for Orthogonal Annotation}

\label{subsec:orthoanno}
%We propose a high-efficient annotation method called orthogonal annotation which only labels two slices in two orthogonal directions for a volumetric image. We now validate our motivation about orthogonal annotation in scarce-supervised segmentation. Our analysis includes the discussion about 1) disparity preservation, 2) informative feature representation, and 3) initial performance promotion. Note that, we take LA dataset~\cite{xiong2021global} as an example, which contains 100 3D images with the detail information listed in Sec.~\ref{subsec:datasets}. For LA dataset, we label a slice in transverse plane and a slice in coronal plane respectively for each volume.

Our proposed orthogonal annotation (OA in short), which only labels two slices in orthogonal directions for a 3D volume, is highly efficient in annotation. Additionally, we now validate its effectiveness empirically and experimentally, which includes the discussion about 1) disparity, 2) feature representation, and 3) initial performance. We here take LA dataset~\cite{xiong2021global} as an example, which contains 100 3D volumes detailed in Sec.~\ref{subsec:datasets}.

%\textbf{\textit{Disparity preservation.}} 

\textbf{OA preserves the disparity.} As revealed by previous theoretical studies in semi-supervised learning~\cite{qiao2018deep}, disparity (or disagreement) between views (or models) providing complementary information is the key to success. %We first disclose the disparity between parallel and orthogonal slices from the perspectives of both image and feature.
Now we explore the similarity of parallel slices and orthogonal slices from both image- and feature-level. The slices are chosen at random and the features are extracted from V-Net~\cite{milletari2016v} trained with fully annotated data.
\begin{wrapfigure}{r}{0.2\textwidth}
\vspace{-20pt}
\begin{center}
\includegraphics[width=0.2\textwidth]{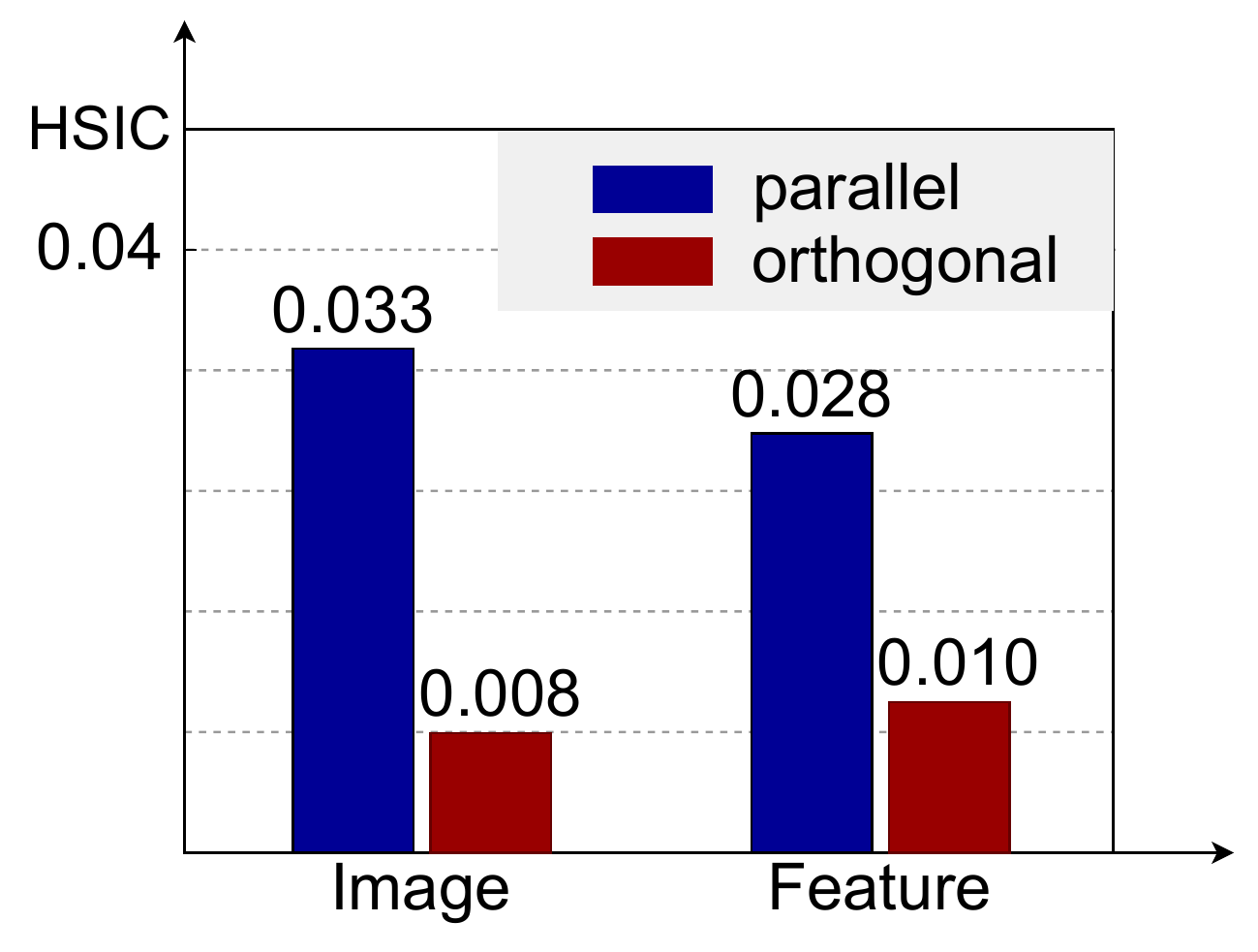}
\end{center}
\vspace{-20pt}

\caption{HSIC of parallel slices and orthogonal slices.}
\label{fig:hsic}
\vspace{-10pt}
\end{wrapfigure}
We utilize the popularly-used Hilbert-Schmidt independence criterion (HSIC)~\cite{ma2020hsic} as the evaluation metric. Smaller value of HSIC, more independence of two variables. The results are shown in Figure~\ref{fig:hsic}, which reveal that parallel slices are more relevant than orthogonal slices in terms of both image and feature. It implies that parallel annotations have more redundancy of information. Besides, slices from different planes are inherently varied and complementary. Thus, for the same amount of annotation, orthogonal annotation could be superior to parallel annotation. 

\textbf{OA enhances consistency from different directions.} We now visualize the features under different settings. Here are three models trained with only one labeled slice in each volume. The labeled slices are from transverse plane for models $t_1$ and $t_2$ and from coronal plane for model $c$. In other words, labeled slices used to train models $t_1$ and $t_2$ are parallel and those used to train models $t_1$ and $c$ are orthogonal. 
\begin{figure}[thb]
\vspace{-10pt}
    \centering
    \includegraphics[width = 0.43\textwidth]{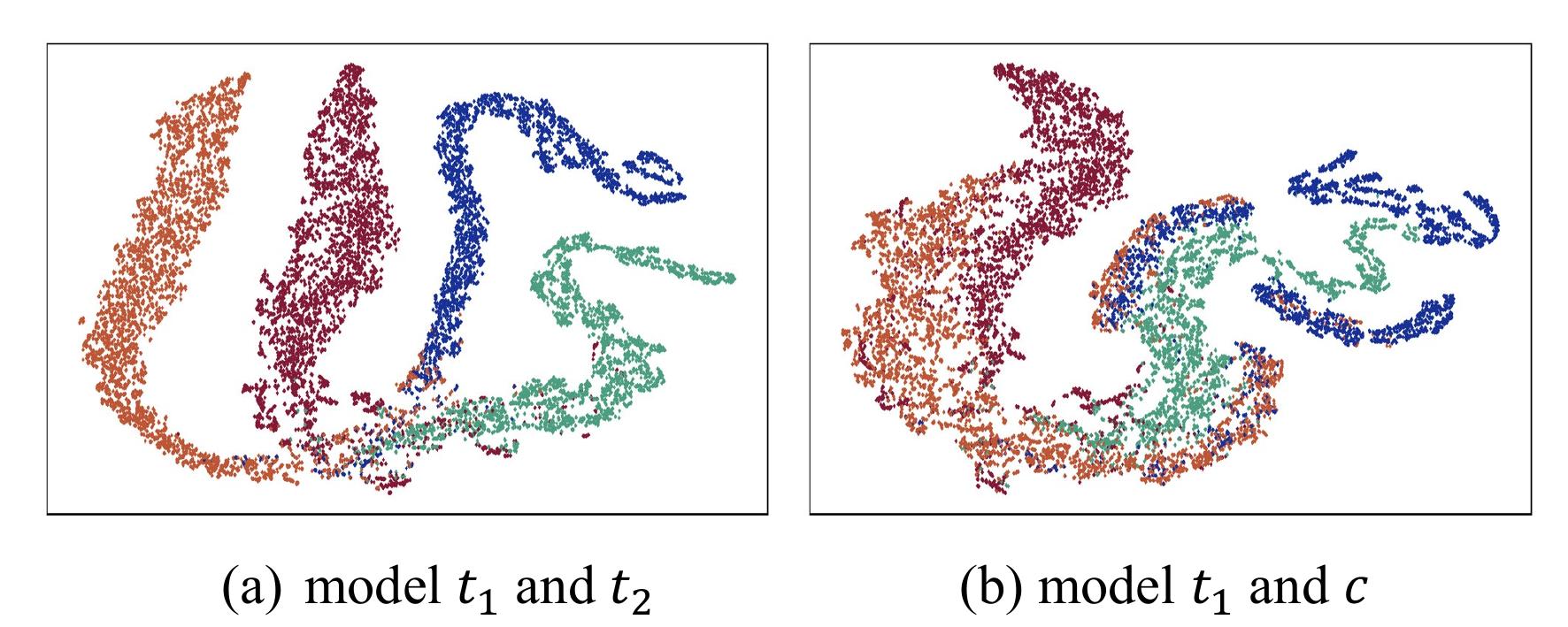}
    \vspace{-1pt}
    \caption{t-SNE result of feature for classification.}
    \label{fig:tsne}
\vspace{-10pt}
\end{figure}

We extract the features before the classification layer of three models and use t-SNE to visualize them. The results are shown in Figure~\ref{fig:tsne}. Red and orange represent positive class while blue and aqua represent negative class. In parallel annotation, even the labeled slices come from a same plane, they learn different sub-patterns to distinguish foreground from background. Differently, the labeled slices to train $t_1$ and $c$ share some common voxels (i.e., overlapping part) so that the features/predictions for the same class are closer. This result validates, by using orthogonal planes, different classes could have more consistent representation.

\textbf{OA provides a promising initialization.} To further explore the utility of the orthogonal annotation, we compare it to the parallel annotation in each plane. We trained models with
three labeling manners: 1) one transverse plane slice and one coronal plane slice 2) two transverse plane slices 3) two coronal plane slices.
\begin{wrapfigure}{l}{0.25\textwidth}
\vspace{-20pt}
\begin{center}
\includegraphics[width=0.25\textwidth]{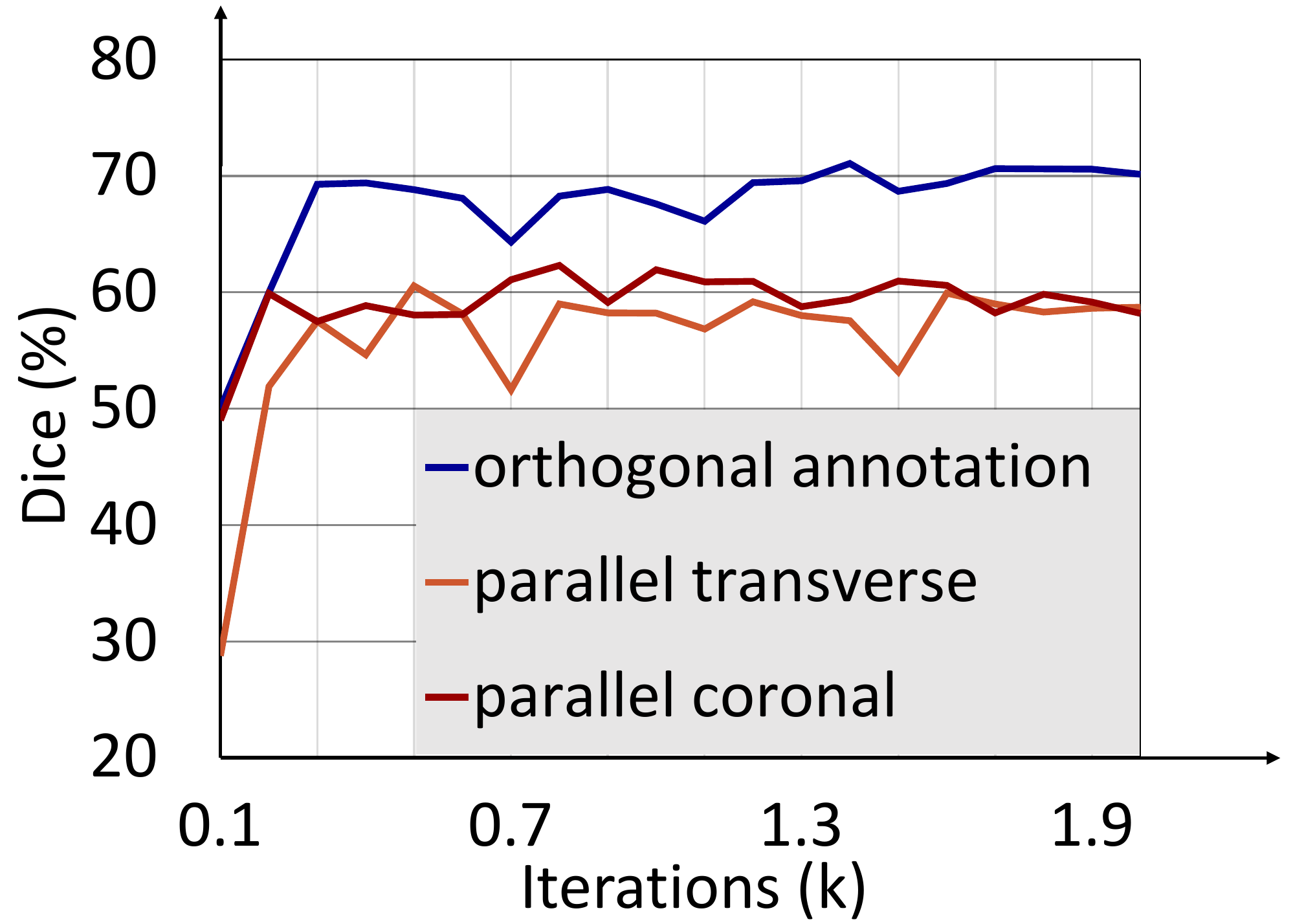}
\end{center}
\vspace{-20pt}
\caption{Dice coefficient on test set of models trained with different annotations.}
\label{fig:laeffe}
\vspace{-10pt}
\end{wrapfigure}
The testing result is shown in Figure~\ref{fig:laeffe}. The model trained with orthogonal annotations steadily and vastly outperform the other two models in its initial stage. It shows that simply leveraging orthogonal plane slices can promote the performance greatly and this fact has always been neglected in medical image segmentation field. 

%These observations have revealed our orthogonal annotation is efficiency in annotation, as well as effectiveness in disparity preservation, feature representation and initial performance promotion. 
These observations have revealed the superiority of our orthogonal annotation. More detailed settings of these experiments are provided in the supplementary material. 

%Based on these observations, we adopt the labeling manner that two orthogonal slices are annotated in a volume.

\subsection{Registration Module}
\label{subsec:registration}
Registration seeks a spatial transformation to map an image to another image. For two slices: labeled slice$(X_l, Y_l)$ and unlabeled slice $X_u$, registration obtains a spatial deformation field $\Phi(\cdot)$ by mapping from $X_l$ to $X_u$. And pseudo label of $X_u$ can be acquired by applying the same spatial deformation field to label $Y_l$, which can be formulated as $\hat Y_u=\Phi(Y_l)$. 

As the segmentation targets in adjacent slices are usually similar in shape and size, it is feasible to propagate the label of a slice to its neighbouring slices through a registration module $\mathcal{M}_{reg}$. Note that registration module is an off-the-shelf tool that requires no training, so it introduces little extra computational cost except generating pseudo labels for only once. 

For volume $X$ with two orthogonal annotated slices $(X_a^m,Y_a^m)$ and $(X_b^n,Y_b^n)$ , we use the two slices to perform slice-by-slice label propagation and gradually generate two pseudo labels $\hat Y_a$ and $\hat Y_b$ for $X$ through registration module $\mathcal{M}_{reg}$. First, we calculate spatial transformation fields that maps $X_a^m$ to its closest slices $X_a^{m-1}$ and $X_a^{m+1}$ through $\mathcal{M}_{reg}$. Then the spatial transformation fields are applied to label $Y_a^m$ to acquire pseudo labels $\hat Y_a^{m-1}$ and $\hat Y_a^{m+1}$, from which we further derive $\hat Y_a^{m-2}$ and $\hat Y_a^{m+2}$. Through the same procedure, we obtain pseudo labels for all slices viewed in plane $a$, and concatenate them to compose pseudo label $\hat Y_a$ for volume $X$. Similarly, the pseudo label $\hat Y_b$ can be easily obtained. So here we have pseudo label sets $\{\hat Y_{1a},\hat Y_{2a},...,\hat Y_{la}\}$ and $\{\hat Y_{1b},\hat Y_{2b},...,\hat Y_{lb}\}$ for those volumes with orthogonal annotation.

Performing label propagation suffers error accumulation as the pseudo label of each unlabeled slice is derived from its nearest pseudo label. Though error propagation can be greatly mitigated through morphology operations, it cannot be completely eliminated due to the limitation of registration itself. To alleviate the problem, we consider the importance of each slice in terms of their credibility.

\subsection{Label Mixing}
For each volume $X$ with sparse orthogonal annotation, we have ground truth sparse annotation $Y$ and dense registration pseudo label $\hat Y_a$ and $\hat Y_b$. Now we generate pseudo label $\tilde Y_a$ and $\tilde Y_b$ by mixing up dense registration pseudo label and ground truth sparse annotation:

\begin{equation}
\tilde Y_a=\textrm{LabelMix}(\hat Y_a,Y),
\end{equation}
\begin{comment}
$$\tilde Y_a^i=\begin{cases}
Y^i \quad\quad \text{if voxel}~i~\text{is labeled in}~Y\\ 
\hat Y_a \quad\quad \text{otherwise,}
\end{cases}$$
\end{comment}
where $\textrm{LabelMix}(\cdot,\cdot)$ denotes a function mixing them by replacing the unlabeled voxels in $Y$ with pseudo label in $\hat Y_a$. Similarly, $\tilde Y_b$ can be generated.

As mentioned in Sec.~\ref{subsec:registration}, different slices in $\hat Y$ should be considered differently according to the quality of pseudo labels. We assume the error rate of label propagation is a certain value, and the credibility of slice $i$ can be denoted as $w_i=\alpha^d$, where $d$ is the distance from slice $i$ to registration source slice and $\alpha$ $(0\le\alpha<1)$ is the decay rate. On these grounds, we set the voxel-wise weight for each voxel in $\tilde Y_a$:
\begin{equation}
 W_a^i=\begin{cases}
1 \quad\quad \text{if voxel}~i~\text{is labeled in}~Y\\ 
\alpha^d \quad\quad \text{otherwise.}
\end{cases}
\end{equation}
Similarly for the voxel weight map of $\tilde Y_b$, defined as $W_b$.
\begin{comment}
W_a^i=\alpha^{d_a\cdot\mathbb{I}(\text{pixel}~i\notin Y)},
\end{comment}
%where $\mathbb{I}(\cdot)$ is an indicator function, 

\subsection{Dense-Sparse Co-Training}
Our DeSCO paradigm consists of two 3D segmentation networks $\mathcal{M}_{seg\cdot a}$ and $\mathcal{M}_{seg\cdot b}$ of a same structure. It leverages labeled and unlabeled volumes simultaneously in a mini-batch. As mentioned above, every volume $X_i$ $(i\leq l)$ has two pseudo labels, ${\tilde Y_{ia}}$ and ${\tilde Y_{ib}}$. 
%Segmentation network $\mathcal{M}_{seg\cdot a}$ regards ${\tilde Y_{ia}}$ as target while $\mathcal{M}_{seg\cdot b}$ regards ${\tilde Y_{ib}}$ as target. 
Segmentation network $\mathcal{M}_{seg\cdot a}$ is trained with ${\tilde Y_{ia}}$ and $\mathcal{M}_{seg\cdot b}$ is trained with ${\tilde Y_{ib}}$, respectively. 
In this way, the two segmentation networks mainly learn from two different perspectives with the targets obtained from different plane registration, and the disparity in orthogonal annotation is well-preserved. The supervised loss contains weighted cross-entropy loss and weighted dice loss, which are formulated as:

\begin{equation}\label{(1)}
\mathcal{L}_{ce}=-\frac{1}{\sum_{i=1}^{H\times W\times D}w_i}\sum_{i=1}^{H\times W\times D}w_iy_i\log p_i,
\end{equation}
and
\begin{equation}\label{(2)}
\mathcal{L}_{dice}=1-\frac{2\times\sum_{i=1}^{H\times W\times D}w_ip_iy_i}{\sum_{i=1}^{H\times W\times D}w_i (p_i^2+ y_i^2)},
\end{equation}
where $w_i$ is $i^{th}$ voxel of weight map $W$ and $p_i$, $y_i$ denote the probability of foreground and pseudo label of voxel $i$ respectively. 

At the early stage of training, segmentation models need to learn from dense pseudo label for steady improvement. Segmentation networks are gradually improved during training and can produce better segmentation results than the initial pseudo labels. Therefore, the initial pseudo labels are actually becoming the obstacle for continual improvement. Based on this fact, we should weaken the influence of pseudo label. To implement this, we gradually decrease the decay rate $\alpha$, and down to zero at last, which means the networks get rid of pseudo labels and only learn from sparse annotation. 

The supervised loss is the weighted sum of cross-entropy loss and dice loss:
\begin{equation}\label{(3)}
\mathcal{L}_{sup}=\frac{1}{2}\mathcal{L}_{ce}+\frac{1}{2}\mathcal{L}_{dice}.
\end{equation}

For those volumes without annotations, the two segmentation models teach each other with their predictions. In a training iteration of volume $X_i$ $(i\geq l+1)$, $\mathcal{M}_{seg\cdot a}$ takes $X_i$ as input and generates its one-hot prediction, which will be partially selected as the pseudo label for $\mathcal{M}_{seg\cdot b}$. Here we follow UAMT~\cite{yu2019uncertainty} by selecting those voxels whose uncertainty is lower than a threshold and produce mask $M_{un}$ for better cross-supervision. Similarly, $\mathcal{M}_{seg\cdot b}$ regards masked one-hot prediction of $\mathcal{M}_{seg\cdot a}$ as the pseudo label. The loss is formulated as:
\begin{equation}\label{(4)}
    \mathcal{L}_{cross}=-\frac{1}{\sum_{i=1}^{H\times W\times D}m_i}\sum_{i=1}^{H\times W\times D}m_i\hat y_i\log p_i,
\end{equation}
where $m_i$ denotes the value in mask $M_{un}$, $p_i$ and $\hat{y_i}$ denote the probability of foreground and one-hot pseudo label predicted by two models.

 For the late stage of training, the supervised loss mainly comes from the labeled slices and the information have been learnt by segmentation models, so keeping supervised loss weight high is to no avail. Oppositely, the cross-supervision weight should be increased for the two networks can correct the mistakes from noisy pseudo label and reach a consensus through cross-supervision. Thus, the overall objective is the weighted sum of $\mathcal{L}_{sup}$ and $\mathcal{L}_{cross}$:

\begin{equation}\label{(5)}
\mathcal{L}=(1-\lambda)\mathcal{L}_{sup}+\lambda\mathcal{L}_{cross},
\end{equation}
where $\lambda$ is a dynamic parameter gradually increasing to optimal cross-supervision weight $\lambda_{oc}$.

\begin{table*}[thb]
\vspace{-10pt}
\renewcommand\arraystretch{0.9}
\vspace{-10pt}
\caption{Comparison on LA dataset segmentation.}
\vspace{-10pt}
\label{tab:la_result}
\centering
\begin{threeparttable}
\resizebox{0.9\linewidth}{!}{
\begin{tabular}{cc|c|c|cc|cccc}
\noalign{\smallskip}
\hline
\noalign{\smallskip}
\multicolumn{2}{c|}{\multirow{2}{*}{Method}} &{\multirow{2}{*}{Venue}}                      &   {\multirow{2}{*}{Setting}}  & \multicolumn{2}{c|}{Scans Used}                           & \multicolumn{4}{c}{Metrics}           \\ 
\cmidrule(r){5-6} \cmidrule(r){7-10}
\multicolumn{2}{c|}{}         & {} && \multicolumn{1}{c|}{\begin{tabular}[c]{@{}c@{}}L / U\\ Volumes\end{tabular}} & Labeled Slices   & Dice (\%)      & Jaccard (\%)   & HD (voxel)   & ASD (voxel)   \\ 
\noalign{\smallskip}
\hline
\noalign{\smallskip}
\multicolumn{1}{c|}{\multirow{7}{*}{Barely-supervised}} & 
\multirow{2}{*}{MT~\cite{tarvainen2017mean}} &\multirow{2}{*}{NIPS'17}& Dense & \multicolumn{1}{c|}{5 / 75}& 10
& 74.41$\pm$4.07 & 60.87$\pm$4.35 &27.87$\pm$3.96 & 5.65$\pm$1.12  \\ 
\multicolumn{1}{c|}{}& {} & &Sparse &\multicolumn{1}{c|}{5 / 75}& 10
& 63.17$\pm$2.28 & 47.48$\pm$2.38 & 39.87$\pm$4.32 & 14.75$\pm$2.41{~~}  \\
\multicolumn{1}{c|}{}& \multirow{2}{*}{CPS~\cite{chen2021semi}}  &\multirow{2}{*}{CVPR'21}& Dense & \multicolumn{1}{c|}{5 / 75}& 10 
& 74.02$\pm$0.41 & 60.62$\pm$1.37 & 30.32$\pm$1.73& 5.86$\pm$0.92  \\ 
\multicolumn{1}{c|}{}& {} && Sparse & \multicolumn{1}{c|}{5 / 75} & 10
& 62.24$\pm$2.55 & 46.12$\pm$2.87 & 45.03$\pm$2.77 & 17.83$\pm$2.29{~~}  \\ 
\multicolumn{1}{c|}{}&\multirow{2}{*}{CTBCT~\cite{luo2021semi1}}  &\multirow{2}{*}{MIDL'22}& Dense &\multicolumn{1}{c|}{5 / 75}& 10 
& 73.12$\pm$5.50 & 59.37$\pm$5.80 & 26.51$\pm$2.47 & \textbf{5.38$\pm$1.25}  \\ 
\multicolumn{1}{c|}{} & {}& & Sparse& \multicolumn{1}{c|}{5 / 75}& 10
& 64.71$\pm$1.49 & 48.89$\pm$1.14 & 42.09$\pm$2.13 & 15.91$\pm$0.83{~~}  \\ 
\multicolumn{1}{c|}{}& \multirow{2}{*}{CoraNet~\cite{shi2021inconsistency}}  &\multirow{2}{*}{TMI'22}& Dense & \multicolumn{1}{c|}{5 / 75} & 10           
&40.63$\pm$1.34 &31.94$\pm$2.41  &40.88$\pm$3.73 &20.19$\pm$1.57{~~~}\\
\multicolumn{1}{c|}{}& &{} & Sparse & \multicolumn{1}{c|}{5 / 75} & 10           
&51.16$\pm$3.63 &38.45$\pm$3.30  &44.70$\pm$1.08  &18.46$\pm$0.78{~~} \\ 
\multicolumn{1}{c|}{}& \textbf{Ours} &\textbf{this paper}& {-} & \multicolumn{1}{c|}{5 / 75} & 10           
&\textbf{77.48$\pm$0.58} & \textbf{64.36$\pm$1.25} & \textbf{{~~}24.86$\pm$10.58} &6.41$\pm$4.68 \\ 
\noalign{\smallskip}
\hline
\noalign{\smallskip}
\multicolumn{1}{c|}{{\color[HTML]{656565}Semi-supervised}}                    & {\color[HTML]{656565}MT~\cite{tarvainen2017mean}}&\color[HTML]{656565}{NIPS'17}& {\color[HTML]{656565}-}       & \multicolumn{1}{c|}{{\color[HTML]{656565}5 / 75}} & {\color[HTML]{656565}440}           & {\color[HTML]{656565}79.13$\pm$2.37}          & {\color[HTML]{656565}66.92$\pm$2.82}          & {\color[HTML]{656565}17.41$\pm$1.85}          & {\color[HTML]{656565}4.76$\pm$1.36}          \\ 
\noalign{\smallskip}
\hline
\noalign{\smallskip}
\end{tabular}}
\end{threeparttable}
\end{table*}

\begin{figure*}[thb]
\vspace{-10pt}
    \centering
    \includegraphics[width = 0.9\textwidth]{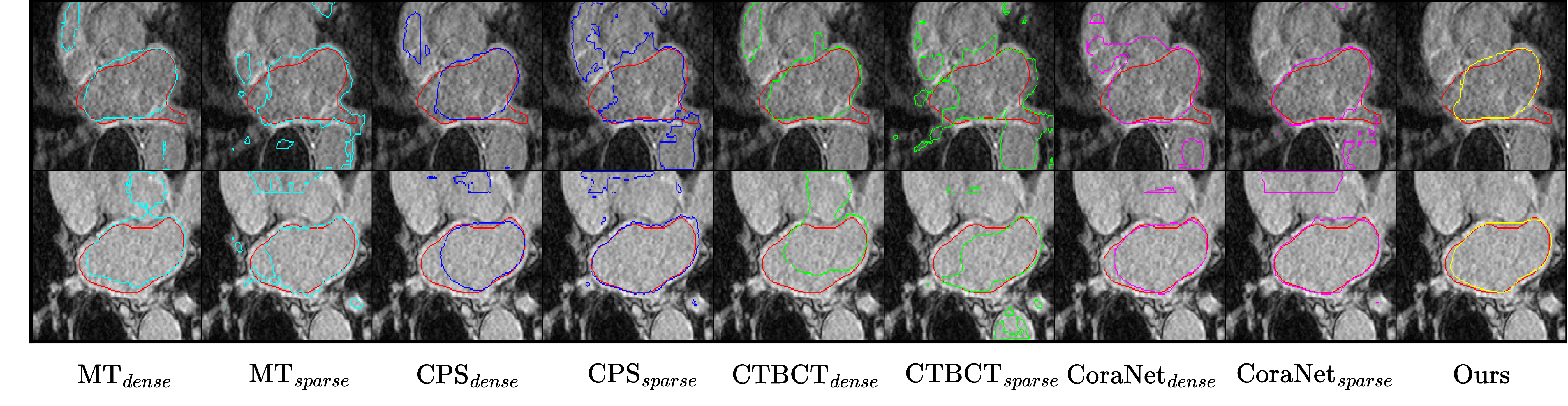}
    \vspace{-10pt}
    \caption{Visual segmentation examples from LA dataset. The red, cyan, blue, green, magenta, yellow curves denote the corresponding results of ground-truth, mean teacher~\cite{tarvainen2017mean}, CPS~\cite{chen2021semi}, CTBCT~\cite{luo2021semi1}, CoraNet~\cite{shi2021inconsistency} and ours, respectively.}
    \label{fig:la_result}
\end{figure*}

\begin{table*}[!h]

\renewcommand\arraystretch{0.9}
\vspace{-10pt}
\caption{Comparison on KiTS19 dataset segmentation.}
\vspace{-10pt}
\label{tab:kits_result}
\centering
\begin{threeparttable}
\resizebox{0.9\linewidth}{!}{
\begin{tabular}{cc|c|c|cc|cccc}
\noalign{\smallskip}
\hline
\noalign{\smallskip}
\multicolumn{2}{c|}{\multirow{2}{*}{Method}} &{\multirow{2}{*}{Venue}}                      &   {\multirow{2}{*}{Setting}}  & \multicolumn{2}{c|}{Scans Used}                           & \multicolumn{4}{c}{Metrics}           \\ 
\cmidrule(r){5-6} \cmidrule(r){7-10}
\multicolumn{2}{c|}{}         & {} && \multicolumn{1}{c|}{\begin{tabular}[c]{@{}c@{}}L / U\\ Volumes\end{tabular}} & Labeled Slices   & Dice (\%)      & Jaccard (\%)   & HD (voxel)   & ASD (voxel)   \\ 
\noalign{\smallskip}
\hline
\noalign{\smallskip}
\multicolumn{1}{c|}{\multirow{7}{*}{Barely-supervised}} & 
\multirow{2}{*}{MT~\cite{tarvainen2017mean}}  &\multirow{2}{*}{NIPS'17}& Dense & \multicolumn{1}{c|}{5 / 185}& 10
 &78.16$\pm$1.99 & 66.96$\pm$1.41 & 17.15$\pm$5.93 & 5.40$\pm$2.25  \\ 
\multicolumn{1}{c|}{}& {} && Sparse &\multicolumn{1}{c|}{5 / 185}& 10
& {~~}32.16$\pm$13.78 & {~~}20.41$\pm$11.04   &{~~}67.42$\pm$12.73  & 32.35$\pm$9.83{~~}  \\ 
\multicolumn{1}{c|}{}& \multirow{2}{*}{CPS~\cite{chen2021semi}}  &\multirow{2}{*}{CVPR'21}& Dense & \multicolumn{1}{c|}{5 / 185}& 10  
& 76.51$\pm$5.81 & 64.67$\pm$6.69 & 17.71$\pm$8.67 & 5.74$\pm$3.50  \\
\multicolumn{1}{c|}{}& {} && Sparse & \multicolumn{1}{c|}{5 / 185} & 10
& {~~}8.87$\pm$2.87 & {~~}4.68$\pm$1.58 & {~~}81.60$\pm$10.15 & 41.21$\pm$5.38{~~}  \\ 
\multicolumn{1}{c|}{}&\multirow{2}{*}{CTBCT~\cite{luo2021semi1}}  &\multirow{2}{*}{MIDL'22}& Dense &\multicolumn{1}{c|}{5 / 185}& 10 
& 80.65$\pm$0.83 & 69.99$\pm$2.27 & 14.88$\pm$6.07 & 4.28$\pm$1.88  \\
\multicolumn{1}{c|}{} & {} & &Sparse& \multicolumn{1}{c|}{5 / 185}& 10
& 17.87$\pm$0.87 & {~~}9.88$\pm$0.55 & 76.06$\pm$8.57& 37.83$\pm$5.17{~~}  \\ 
\multicolumn{1}{c|}{}& \multirow{2}{*}{CoraNet~\cite{shi2021inconsistency}}  &\multirow{2}{*}{TMI'22}& Dense & \multicolumn{1}{c|}{5 / 185} & 10           
&{~~}43.08$\pm$28.61 & {~~}33.27$\pm$23.87 &24.17$\pm$3.88  &8.61$\pm$2.11 \\ 
\multicolumn{1}{c|}{}& {} && Sparse & \multicolumn{1}{c|}{5 / 185} & 10           
&{~~}28.64$\pm$25.05 & {~~}18.90$\pm$19.44 &{~~}72.62$\pm$22.57  &35.22$\pm$13.43 \\ 
\multicolumn{1}{c|}{}& \textbf{Ours} &\textbf{this paper}& {-} & \multicolumn{1}{c|}{5 / 185} & 10           
& \textbf{86.93$\pm$2.78} & \textbf{78.33$\pm$3.21} & \textbf{11.61$\pm$2.15}   & \textbf{3.28$\pm$0.51}  \\ 
\noalign{\smallskip}
\hline
\noalign{\smallskip}
\multicolumn{1}{c|}{{\color[HTML]{656565}Semi-supervised}}                    & {\color[HTML]{656565}MT~\cite{tarvainen2017mean}} & {\color[HTML]{656565}NIPS'17}& {\color[HTML]{656565}-}       & \multicolumn{1}{c|}{{\color[HTML]{656565}5 / 185}}                                                              & {\color[HTML]{656565}320}           & {\color[HTML]{656565}84.98$\pm$4.68}          & {\color[HTML]{656565}76.51$\pm$5.63}          & {\color[HTML]{656565}16.51$\pm$0.18}          & {\color[HTML]{656565}4.61$\pm$0.83}          \\ 
\noalign{\smallskip} 
\hline
\noalign{\smallskip}

\end{tabular}}
\end{threeparttable}
\end{table*}
\begin{figure*}[!h]
\vspace{-10pt}
    \centering
    \includegraphics[width = 0.9\textwidth]{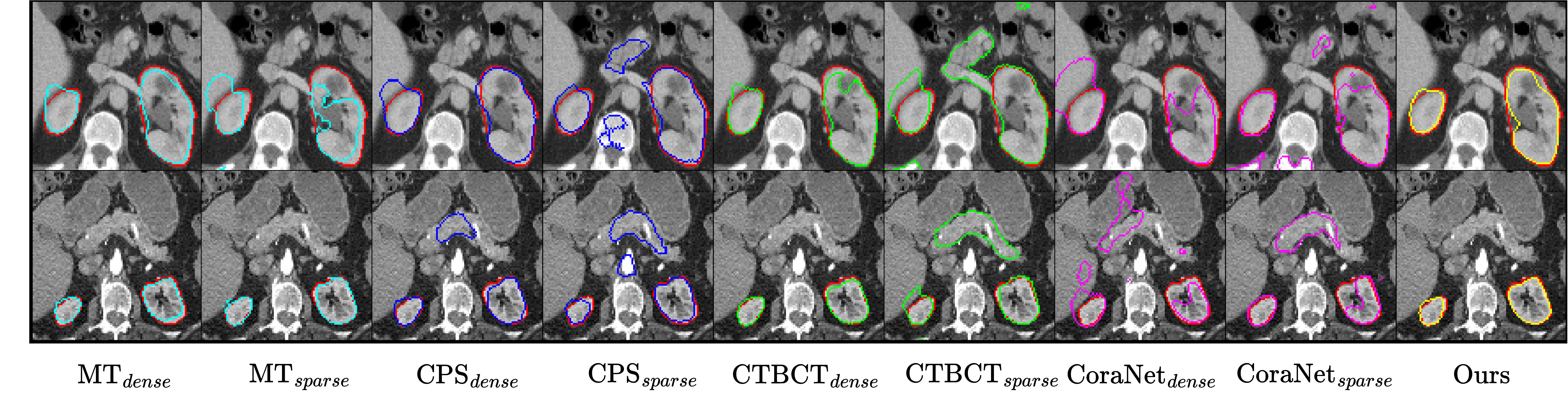}
    \vspace{-10pt}
    \caption{Visual segmentation examples from KiTS19 dataset. The red, cyan, blue, green, magenta, yellow curves denote the corresponding results of ground-truth, mean teacher~\cite{tarvainen2017mean}, CPS~\cite{chen2021semi}, CTBCT~\cite{luo2021semi1}, CoraNet~\cite{shi2021inconsistency} and ours, respectively.}
    \vspace{-10pt}
    \label{fig:kits_result}
\end{figure*}

%preprocessing write in here
\section{Experiments}
\label{sec:Experiments}
\subsection{Datasets}
\label{subsec:datasets}

%\begin{itemize}
%    \item \textbf{LA Dataset}~\cite{xiong2021global} contains 100 3D gadolinium-enhanced magnetic resonance images required using clinical whole-body MRI scanner and full annotations for left atrial cavity by radiologists. The raw MRIs are gray-scale images and the labels are binary. All scans have the same isotropic resolution of resolution of $0.625\times0.625\times0.625$mm$^3$ while their dimensions vary from each other. 
%    \item \textbf{KiTS19 Dataset}~\cite{heller2019kits19} is a popularly used benchmark for kidney and kidney tumor segmentation. It contains 300 arterial phase abdominal CT where there are one or more kidney tumors and 210 of them have accessible kidney and kidney tumor delineations given by medical students under the supervision of radiologists. The CTs have an average slices number of 216 and the slices thicknesses range from 1mm to 5mm. Most of slices in the transverse plane have the size of $512 \times 512$.
%    \item \textbf{LiTS Dataset}~\cite{bilic2019liver} consists of 201 contrast-enhanced abdominal CT scans and corresponding segmentation of liver as well as liver tumor provided by various clinical sites around the world. And 131 of segmentations are open to the public. There are high variations in pixel spacing and slice thickness.
%\end{itemize}

\begin{table*}[thb]
\vspace{-10pt}
\renewcommand\arraystretch{0.9}
\caption{Comparison on LiTS dataset segmentation.}
\vspace{-10pt}
\label{tab:lits_result}
\centering
\begin{threeparttable}
\resizebox{0.9\linewidth}{!}{
\begin{tabular}{cc|c|c|cc|cccc}
\noalign{\smallskip}
\hline
\noalign{\smallskip}
\multicolumn{2}{c|}{\multirow{2}{*}{Method}} &{\multirow{2}{*}{Venue}}                      &   {\multirow{2}{*}{Setting}}  & \multicolumn{2}{c|}{Scans Used}                           & \multicolumn{4}{c}{Metrics}           \\ 
\cmidrule(r){5-6} \cmidrule(r){7-10}
\multicolumn{2}{c|}{}         & {} && \multicolumn{1}{c|}{\begin{tabular}[c]{@{}c@{}}L / U\\ Volumes\end{tabular}} & Labeled Slices   & Dice (\%)      & Jaccard (\%)   & HD (voxel)   & ASD (voxel)   \\ 
\noalign{\smallskip}
\hline
\noalign{\smallskip}
\multicolumn{1}{c|}{\multirow{7}{*}{Barely-supervised}} & 
\multirow{2}{*}{MT~\cite{tarvainen2017mean}} &\multirow{2}{*}{NIPS'17}& Dense & \multicolumn{1}{c|}{5 / 95}& 10
 & 81.76$\pm$4.82 & 69.73$\pm$7.01 & {~~}28.87$\pm$13.85 & 8.57$\pm$4.03  \\ 
\multicolumn{1}{c|}{}& {} & &Sparse &\multicolumn{1}{c|}{5 / 95}& 10
& {~~}56.82$\pm$25.76 & {~~}43.83$\pm$28.78 & {~~}74.03$\pm$34.02 & 31.41$\pm$17.58  \\  
\multicolumn{1}{c|}{}& \multirow{2}{*}{CPS~\cite{chen2021semi}} &\multirow{2}{*}{CVPR'21}& Dense & \multicolumn{1}{c|}{5 / 95}& 10   
& {~~}73.86$\pm$10.13 &{~~}59.73$\pm$13.16 & {~~}32.78$\pm$14.68 & 10.73$\pm$5.20{~~}  \\ 
\multicolumn{1}{c|}{}& {} & &Sparse & \multicolumn{1}{c|}{5 / 95} & 10
 & 20.46$\pm$2.15 & 11.48$\pm$1.35 & 92.27$\pm$3.03 & 41.34$\pm$2.09{~~}  \\ 
\multicolumn{1}{c|}{}&\multirow{2}{*}{CTBCT~\cite{luo2021semi1}} &\multirow{2}{*}{MIDL'22}& Dense &\multicolumn{1}{c|}{5 / 95}& 10 
& 79.68$\pm$5.45 & 66.95$\pm$7.41 & {~~}30.46$\pm$13.36 & 9.18$\pm$4.12  \\ 
\multicolumn{1}{c|}{} & {} & &Sparse& \multicolumn{1}{c|}{5 / 95}& 10
& 40.07$\pm$7.95 & 25.52$\pm$6.24 & {~~}71.83$\pm$11.41 & 29.49$\pm$5.44{~~}  \\ 
\multicolumn{1}{c|}{}& \multirow{2}{*}{CoraNet~\cite{shi2021inconsistency}}&\multirow{2}{*}{TMI'22} & Dense & \multicolumn{1}{c|}{5 / 95} &  10          
& 80.17$\pm$1.77& 68.97$\pm$3.16 &19.42$\pm$7.47  &4.34$\pm$0.50 \\ 
\multicolumn{1}{c|}{}& {} & &Sparse & \multicolumn{1}{c|}{5 / 95} &  10          
& 36.84$\pm$8.20& 23.13$\pm$6.03 &95.89$\pm$1.94  &43.25$\pm$2.03{~~} \\ 
\multicolumn{1}{c|}{}& \textbf{Ours} & \textbf{this paper}&{-} & \multicolumn{1}{c|}{5 / 95} & 10           
& \textbf{89.24$\pm$1.37} & \textbf{81.10$\pm$2.28} & \textbf{10.05$\pm$2.42} & \textbf{2.27$\pm$0.45}  \\ 

\noalign{\smallskip}
\hline
\noalign{\smallskip}
\multicolumn{1}{c|}{{\color[HTML]{656565}Semi-supervised}}                    & {\color[HTML]{656565}MT~\cite{tarvainen2017mean}} &\color[HTML]{656565}{NIPS'17}& {\color[HTML]{656565}-}       & \multicolumn{1}{c|}{{\color[HTML]{656565}5 / 95}}                                                              & {\color[HTML]{656565}320}           & {\color[HTML]{656565}90.77$\pm$1.91}          & {\color[HTML]{656565}83.62$\pm$3.11}          & {\color[HTML]{656565}18.32$\pm$7.17}          & {\color[HTML]{656565}4.95$\pm$1.81}          \\ 
\noalign{\smallskip}
\noalign{\smallskip}
\hline
\end{tabular}}
\end{threeparttable}
\end{table*}
\begin{figure*}[thb]
\vspace{-10pt}
    \centering
    \includegraphics[width = 0.9\textwidth]{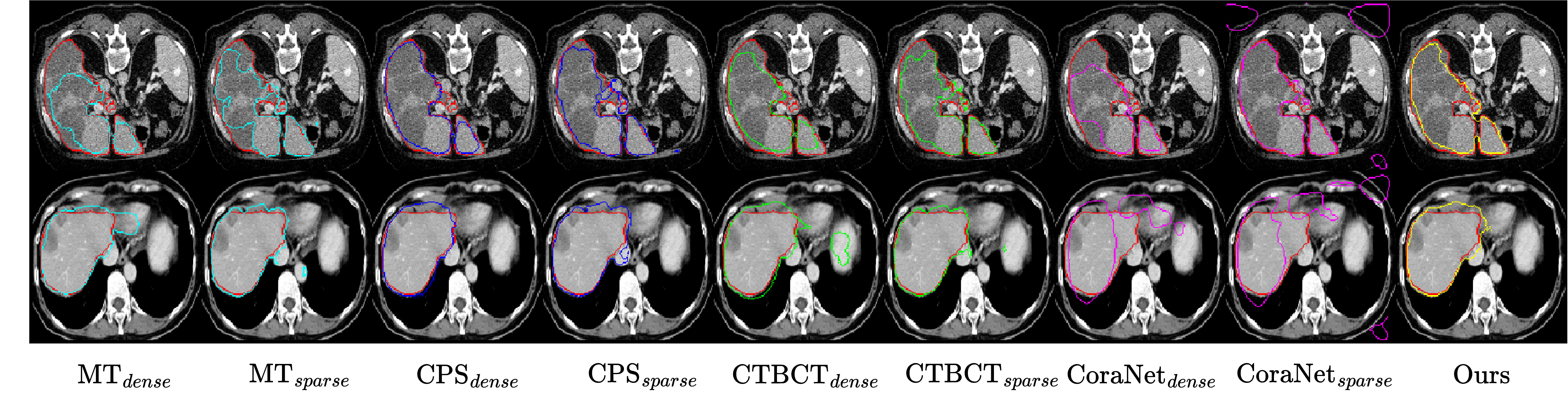}
    \vspace{-10pt}
    \caption{Visual segmentation examples from LiTS dataset. The red, cyan, blue, green, magenta, yellow curves denote the corresponding results of ground-truth, mean teacher~\cite{tarvainen2017mean}, CPS~\cite{chen2021semi}, CTBCT~\cite{luo2021semi1}, CoraNet~\cite{shi2021inconsistency} and ours, respectively.}
    \label{fig:lits_result}
\end{figure*}
\textbf{LA Dataset}~\cite{xiong2021global} contains 100 3D gadolinium-enhanced magnetic resonance images required using clinical whole-body MRI scanner and full annotations for left atrial cavity by radiologists. All scans have the same isotropic resolution of resolution of $0.625\times0.625\times0.625$mm$^3$ while their dimensions vary from each other. 
 
\textbf{KiTS19 Dataset}~\cite{heller2019kits19} is a benchmark for kidney and kidney tumor segmentation. It contains 300 arterial phase abdominal CT where there are one or more kidney tumors and 210 of them have accessible kidney and kidney tumor delineations. The CTs have an average slice number of 216 and the slices thicknesses range from 1mm to 5mm. Most of slices in the transverse plane have the size of $512 \times 512$.

\textbf{LiTS Dataset}~\cite{bilic2019liver} consists of 201 contrast-enhanced abdominal CT scans and corresponding segmentation of liver as well as liver tumor provided by various clinical sites. And 131 of segmentations are open to the public. There are high variations in pixel spacing and slice thickness.

\subsection{Experiment Setting}
For the selection of annotated slices, the principle is that the segmented target should be visible in the selected slice. For generalization, we simply select a fixed slice roughly located in the middle of the whole target.

For the implementation of our method, we adopt the SyNRA method in Ants~\cite{avants2011reproducible} as our registration module and V-Net~\cite{milletari2016v} as the backbone of our segmentation models. 

For the training of segmentation models, we set the batch size as 2, one for labeled sample and one for unlabeled sample. We use Stochastic Gradient Descent (SGD) with momentum of 0.9 and weight decay of 0.0001 as optimizer. The learning rate is initialized as 0.01 and gradually decays to 0.0001. Training iteration is set to 6000 for all experiments. $\lambda_{oc}$ is 0.8 for all three datasets. For the slice-wise weight decay rate, we initialize $\alpha$ to 0.95 and update every 1000 iterations according to cosine rampdown from~\cite{loshchilov2016sgdr}.

For the evaluation and comparison of our method and other semi-supervised methods, we adopt four commonly used metrics, which are Dice coefficient, Jaccard coefficient, 95\% Hausdorff Distance (95\% HD) and Average Surface Distance (ASD). We compare our method with some classic and state-of-the-art (SOTA) methods, including Mean Teacher (\textbf{MT})~\cite{tarvainen2017mean}, Cross Pseudo Supervision (\textbf{CPS})~\cite{chen2021semi}, Cross Teaching Between CNN and Transformer (\textbf{CTBCT})~\cite{luo2021semi1} and Conservative-radical Network (\textbf{CoraNet})~\cite{shi2021inconsistency}. All methods are implemented in 3D manners. MT, CPS, CoraNet and the CNN network of CTBCT use V-Net~\cite{milletari2016v} as backbone, the transformer network of CTBCT is implemented as UNETR, which is introduced in~\cite{hatamizadeh2022unetr}. As those methods are not specifically designed for barely-supervised setting, for fairer comparisons, we designed two training settings for those methods. 1) Dense: we provided dense pseudo labels by registration as labeled training data. From the comparison between our method and other methods trained in this setting, we show that our method can better leverage the inaccurate pseudo labels while other methods suffers severe degradation. 2) Sparse: we provide original orthogonal labels without registration. And during training, the supervised loss only comes from labeled voxles. In the comparison between our method and other methods trained in this setting, we show that directly learning from few sparse annotations is almost unattainable, thus, generating initial pseudo labels is necessary and reasonable. 
All the experiments are implemented using PyTorch and an NVIDIA GeForce RTX 3090 GPU.

%data split and dataset specific operation write in here.
\subsection{Compared with SOTA Methods}
\textbf{Results on LA Dataset.} Following~\cite{yu2019uncertainty}, we use 80 volumes for training and 20 for testing. 5 volumes of training data have orthogonal annotations on transverse plane and coronal plane, while the rest of the training data are used as unlabeled volumes. As there's no validation set in the experiments, so we report the results at the end of 6000 iterations of training for all methods. We randomly crop the original size $112\times 112 \times 88$ into $112\times 112\times 80$ for training, and the patch size remains the same during testing. The prediction is generated by sliding the patch window with a stride of 18 on the coronal and sagittal plane and 4 on the transverse plane. The comparison results between our method and other methods are presented as mean $\pm$ std of five cross-validation in Table~\ref{tab:la_result}. It is clear that our method outperforms other SOTA semi-supervised methods in all metrics except ASD. Compared to Mean Teacher trained with fully annotated volumes, our method uses only 2.3\% annotations but there is only a 1.65\% drop on Dice coefficient. We notice that these methods in Sparse setting work poorly, it is mainly because of the variety between slices and the lack of supervision signal on most parts of the whole volume. The segmentation result examples are shown in Figure~\ref{fig:la_result}.

\textbf{Results on KiTS19 Dataset.}
We divided 210 samples into training set and test set, which have 190 and 20 samples respectively. Similar to LA, 5 volumes are randomly selected to label and the rest 185 volumes are unlabeled data. Patch size is set to $128\times 128\times 64$. The results are shown in Table~\ref{tab:kits_result}. From the table we can see that our method outperforms all methods on all metrics. Note that our method even outperforms MT trained with 32 times the annotated slices. The segmentation result examples are shown in Figure~\ref{fig:kits_result}.

%\begin{figure*}[thb]
%    \centering
%    \includegraphics[width = 0.94\textwidth]{img/caption.png}
%    
%    \label{fig:lits_result2}
%\end{figure*}
\textbf{Results on LiTS Dataset.}
For LiTS dataset, we also randomly select 5 volumes as labeled data, 95 volumes as unlabeled data and the rest 31 as testing data. We adopt random cropping to get patches of $176\times 176\times 64$ for training and testing. From Table~\ref{tab:lits_result}, our method uses only 10 labeled slices and obtains a Dice coefficient close to 90\%. Compared with other methods, our method outperforms all methods by a large margin. And our method is only 1.53\% lower than MT trained with full volumetric annotation on Dice metric, but performs much better on HD and ASD. The segmentation result examples are shown in Figure~\ref{fig:lits_result}.
\subsection{Ablation Study}
\textbf{Effectiveness of Each Component:}
In order to better understand and evaluate the components of our method, we conduct an ablation experiment on the KiTS19 dataset. Now we introduce each setting of our experiment: 1) training from mixed dense pseudo label (Dense), 2) leveraging slice-wise weight ($w_s$), 3) uncertainty-guided cross-supervision ($M_{un}$), 4) gradually decreasing the weight decay rate $\alpha$ to zero, \ie, learning from dense to sparse ($D\rightarrow S$), and 5) gradually increasing cross-supervision weight to $\lambda_{oc}$ ($\lambda\uparrow$). The results are shown in Table~\ref{tab:ablation}.

\begin{table}[t]
\vspace{-10pt}
\renewcommand\arraystretch{1}
\setlength\tabcolsep{4.5pt}
\caption{Ablation study on KiTS19 dataset.}
\vspace{-10pt}
\label{tab:ablation}
\centering
\resizebox{1\linewidth}{!}{
\begin{tabular}{c|ccccc|cccc}
\noalign{\smallskip}
\hline
\noalign{\smallskip}
Methods &Dense&$w_s$&$M_{un}$&$D\rightarrow S$&$\lambda\uparrow$ & Dice (\%)  & Jaccard (\%)   & HD (voxel)   & ASD (voxel)   \\ 
\noalign{\smallskip}
\hline
\noalign{\smallskip}
\#1 &\checkmark && && &75.33$\pm$8.77 & 63.20$\pm$10.95 & 19.14$\pm$8.14 & 6.53$\pm$3.27  \\ 
\#2 &\checkmark &\checkmark & && &77.69$\pm$5.07 & 65.75$\pm$6.57 & 17.90$\pm$ 6.34 & 5.65$\pm$2.47  \\ 
\#3 &\checkmark &\checkmark &\checkmark && & 78.31$\pm$3.94 & 66.48$\pm$4.67 & 18.35$\pm$5.59 & 5.43$\pm$1.62  \\ 
\#4  & \checkmark&\checkmark &\checkmark &\checkmark& & 78.45$\pm$6.45 & 67.03$\pm$9.13 & 30.50$\pm$6.28 & 9.57$\pm$3.32  \\ 
\#5  & \checkmark&\checkmark &\checkmark &&\checkmark & 84.86$\pm$2.01 & 74.99$\pm$2.16 & 12.21$\pm$2.75 & 3.34$\pm$0.95  \\ 

\noalign{\smallskip}
\hline
\noalign{\smallskip}
\#6 & \checkmark&\checkmark &\checkmark &\checkmark &\checkmark& \textbf{86.93$\pm$2.78} & \textbf{78.33$\pm$3.21} & \textbf{11.61$\pm$2.15}   & \textbf{3.28$\pm$0.51}\\
\noalign{\smallskip}
\hline
\noalign{\smallskip}
\end{tabular}}
\vspace{-10pt}
\end{table}
 The process of learning from dense to sparse and a high cross-supervision weight in the late stage play the most critical role. Throughout the training procedure, the performance of the segmentation models is improving, and they can produce better results than the initial pseudo labels, so the high weighted supervised loss is gradually hampering the segmentation networks from improving. Simply leveraging slice-wise weight $w_s$ is also effective.

\begin{figure}[thb]
\vspace{-10pt}
    \centering
    \includegraphics[width = 0.40\textwidth]{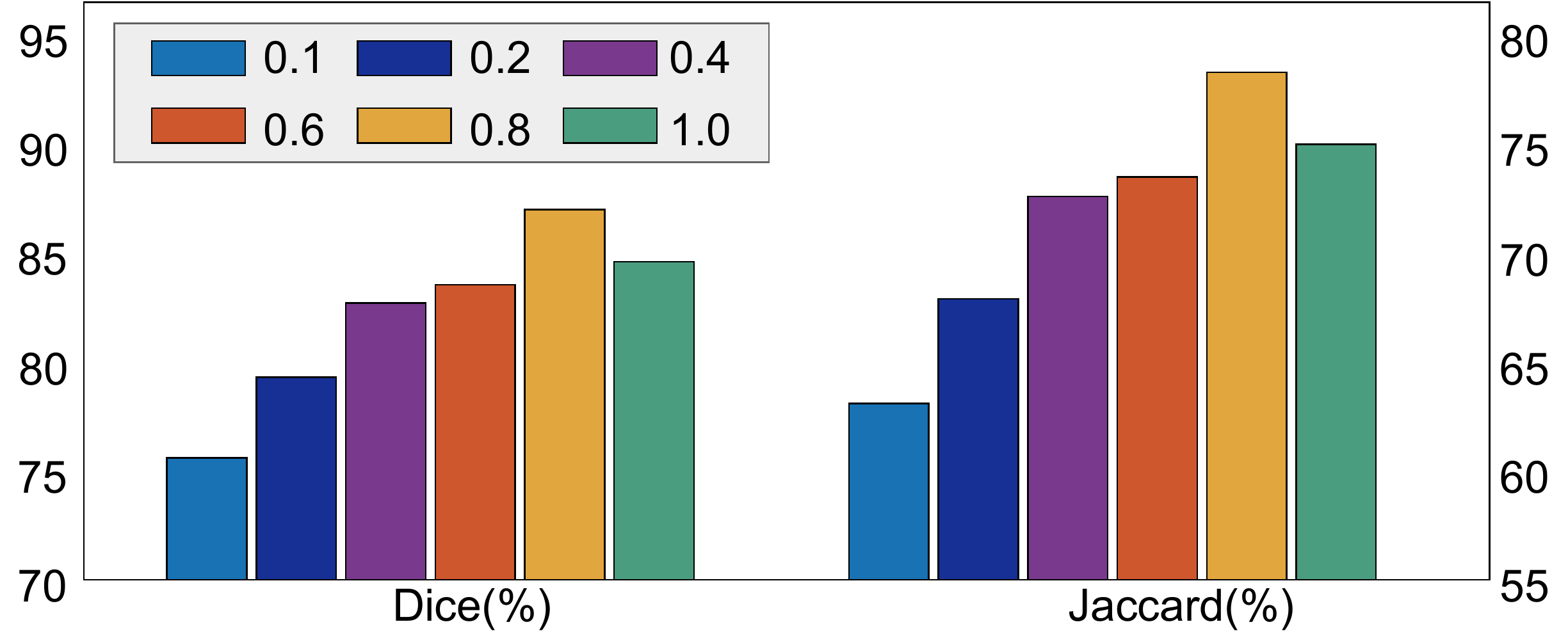}
    \vspace{-10pt}
    \caption{Segmentation result on KiTS dataset with different optimal cross-supervision weight $\lambda_{oc}$.}
    \label{fig:empirical}
    \vspace{-10pt}
\end{figure}

\textbf{Quantitative Analysis on Hyper-Parameter:}
We also conduct exhaustive experiments of different optimal cross-supervision weights $\lambda_{oc}$ on the KiTS19 datasets. Specifically, we set the optimal cross-supervision weight as 0.1, 0.2, 0.4, 0.6, 0.8 and 1.0. We find that setting $\lambda_{oc}$ as 0.8 serves best. Detailed results are shown in Figure~\ref{fig:empirical}.

It is shown that optimal cross-supervision weight should not be too large or too small, because small $\lambda_{oc}$ makes the networks concentrate on learning from registration labels and neglect the fact that the networks are improving during the training procedure, while too large $\lambda_{oc}$ makes the network gradually lose the supervision from the ground truth labels, and the performance tends to decrease through further cross-supervision without constraint.

\section{Conclusion}
In this paper, we proposed orthogonal annotation for 3D medical image segmentation which is to label two orthogonal slices for a volume, and we verified its high-efficiency. Then we applied this annotation manner in barely-supervised segmentation setting. To better leverage the volumes with orthogonal annotation and the large number of unlabeled volumes, we designed a dense-sparse co-training paradigm, which learns from dense pseudo label first, then reduces noise and gains further promotion from sparse annotation in later stage. Our method could better synthesize the knowledge from two planes through cross-supervision. Large quantities of experiments have validated the effectiveness of our orthogonal annotation and DeSCO paradigm, \eg, our method could achieve a close or better performance compared with semi-supervised method, but with only 2\%$\sim$3\% annotations.

%%%%%%%%% REFERENCES
{\small
\bibliographystyle{ieee_fullname}
\bibliography{ref}
}
\section*{A. Algorithm Summary}
The training procedure of our method is summarized in Algorithm \ref{alg:A}.
\begin{algorithm}[thb]
\caption{Training Procedure with Sparse Orthogonal Annotation on Tiny Fraction of Volumes}  
\label{alg:A}  
\LinesNumbered 
For \textcolor{red}{*} steps: repeat at once by exchanging $a$ and $b$.\\
\KwIn{Training images $\{X_i|i\leq N\}$\\
\thinspace Orthogonal annotations $\{Y_i:(Y_{ia}^{mi},Y_{ib}^{ni})|i\leq l\}$}
\KwOut{Segmentation models $\mathcal{M}_{seg\cdot a}$, $\mathcal{M}_{seg\cdot b}$}
\textcolor{blue}{\texttt{//pseudo label generation}}\\
\For{i $\in [1,l]$}{
    \textcolor{red}{*}pseudo label $\hat Y_{ia}\leftarrow \mathcal{M}_{reg}(X_i,Y_{ia}^{mi})$\\
    \textcolor{red}{*}pseudo label $\tilde Y_{ia}\leftarrow \textrm{LabelMix}(\hat Y_{ia},Y_i)$\\
}
\textcolor{blue}{\texttt{//initialization}}\\
Cross-supervision weight $\lambda\leftarrow0$\\
Decay rate $\alpha\leftarrow0.95$\\
Generating weight map $W_{a}$ and $W_{b}$ with $\alpha$\\ 
\textcolor{blue}{\texttt{//model training}}\\
\While{not converged}{
    \ForEach{$X_i$ in minibatch}{
    \textcolor{red}{*}$P_{ia}$, $\bar P_{ia}$, $M_{un\cdot a}$  $\leftarrow \mathcal{M}_{seg\cdot a}(X_i)$\\
    \eIf{$i\leq l$}{
        
        \textcolor{red}{*}$\mathcal{L}_{sup\cdot a}\leftarrow \mathcal{L}_{sup}(P_{ia},\tilde Y_{ia},W_a)$\\
    }{
        
        \textcolor{red}{*}$\mathcal{L}_{cross\cdot a}\leftarrow \mathcal{L}_{cross}(P_{ia},\bar P_{ib},M_{un\cdot b})$\\

    }
    
}
    \textcolor{red}{*}$\mathcal{L}_a\leftarrow (1-\lambda)\mathcal{L}_{sup\cdot a}+\lambda\mathcal{L}_{cross\cdot a}$\\
    
    \textcolor{blue}{\texttt{//dense to sparse}}\\
    Update decay rate $\alpha$, weight map $W_a, W_b$\\
    Update cross-supervision weight $\lambda$\\
    Update $\mathcal{M}_{seg\cdot a}$, $\mathcal{M}_{seg\cdot b}$
}
\end{algorithm}
\label{sec:algorithm}

\section*{B. Detailed Settings of Experiments in Sec. 3.2}
This section provides detailed settings of three experiments in Sec. 3.2 in order.

\textbf{Setting 1.} The V-Net~\cite{milletari2016v} model is trained with 100 fully annotated volumes. The features are extracted from the layer before classification layer. The parallel slices are from transverse plane, the orthogonal slices are from transverse plane and coronal plane. As the input volumes are cropped into $112\times112\times80$, the serial numbers for slices in transverse plane are randomly sampled from [1,80] and the serial numbers for slices in coronal plane are randomly sampled from [1,112], respectively. The results reported in Figure 3 of the main paper are the average HSIC~\cite{ma2020hsic} value of the slices selected from 100 volumes.

\textbf{Setting 2.} We randomly sample three slice serial number $s_{t_1}$, $s_{t_2}$ and $s_{c}$. And the labeled slices to train models $t_1$, $t_2$ and $c$ are the ${s_{t_1}}^{th}$ slice from transverse plane, ${s_{t_2}}^{th}$ slice from transverse plane and ${s_{c}}^{th}$ slice from coronal plane, respectively. The illustration in Figure 4 of the main paper is generated when $s_{t_1}=43$, $s_{t_2}=53$ and $s_{c}=70$. And as long as the slices selected from transverse planes do not separate too far, the property illustrated in Figure 4 still holds.

\textbf{Setting 3.} 
The models in this part are trained with 8 volumes where only two slices are labeled in each volume. The serial numbers for slices in transverse plane are randomly sampled from [1,88], and the serial numbers for slices in coronal plane are randomly sampled from [1,132].

\section*{C. Monitoring of the Dice Coefficient on KiTS19 Dataset}
\begin{figure}[t]
\vspace{-10pt}
    \centering
    \includegraphics[width = 0.45\textwidth]{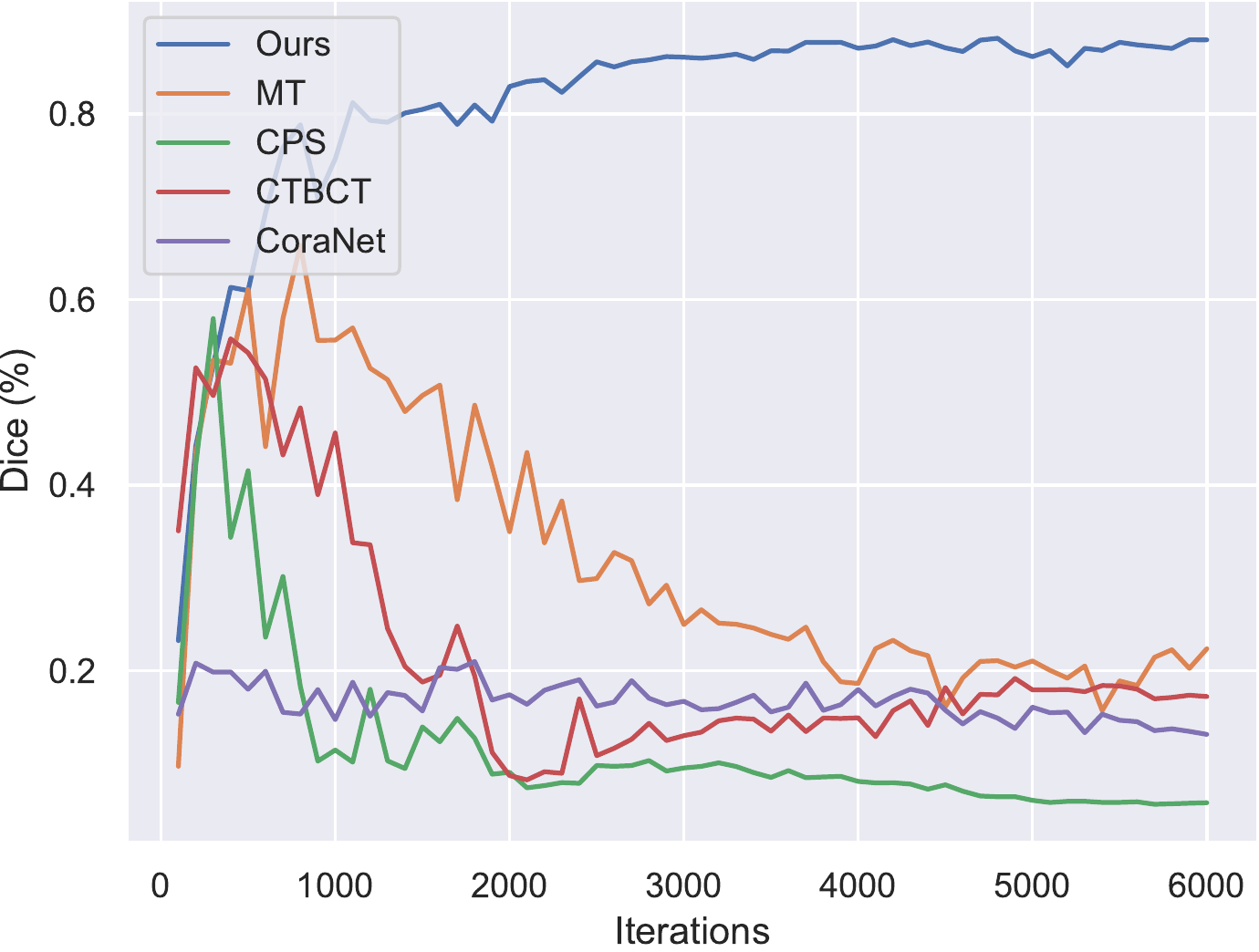}
    \vspace{-10pt}
    \caption{The performance comparison between our method and other methods (MT~\cite{tarvainen2017mean}, CPS~\cite{chen2021semi}, CTBCT~\cite{luo2021semi1}, CoraNet~\cite{shi2021inconsistency}) trained in Sparse setting on KiTS19~\cite{heller2019kits19} dataset.}
    \label{fig:d2s}
    \vspace{-10pt}
\end{figure}

To better illustrate 1) the effectiveness of our method in achieving stable and continuous improvement through the whole training process and 2) the difficulty of directly learning from sparse annotations, we save the intermediate models trained in Sec. 4.3 (MT~\cite{tarvainen2017mean}, CPS~\cite{chen2021semi}, CTBCT~\cite{luo2021semi1}, CoraNet~\cite{shi2021inconsistency} trained in \textbf{Sparse} setting and Ours) every 100 iterations and test their performance. The experiment setting has been introduced in Sec. 4.2. We use KiTS19~\cite{heller2019kits19} dataset as an example and the result is shown in Figure \ref{fig:d2s}.

As shown in Figure \ref{fig:d2s}, our method achieves stable and continuous improvement, while other methods directly learning from sparse annotations suffer extremely unstable training and performance degradation in later stage. Also, their peak performances are inferior to ours.

\end{document}